\documentclass[preprint,12pt]{elsarticle}

\usepackage[colorlinks=true,linkcolor=blue,citecolor=blue,urlcolor=blue]{hyperref}
\usepackage{amsmath,amssymb,amsfonts}
\usepackage{booktabs}
\usepackage{multirow}
\usepackage{graphicx}
\usepackage{xcolor}
\usepackage{algorithm}
\usepackage{algpseudocode}
\usepackage{microtype}
\usepackage{bm}
\usepackage{url}
\usepackage{float} 
 
\journal{Pattern Recognition}
 
\newcommand{\doi}[1]{\href{https://doi.org/#1}{\nolinkurl{doi:#1}}}

\newcommand{\vu}{\bm{u}}
\newcommand{\vn}{\bm{n}}

\begin{document}
 
\begin{frontmatter}
\title{Structural Kolmogorov--Arnold Convolutions: Learnable
       Function on the Values or the Filter Shape as Parameter-Efficient
       Alternative to Per-Edge Convolutional KANs}
 
\author[iit,dibris]{Stefano Mereu}
\author[ecampus,karazin]{Oleksandr Kuznetsov}
\author[iit]{Gabriele Marchello}
\author[univpm]{Alessandro Galdelli}
\author[univpm]{Emanuele Frontoni}
\author[univpm]{Adriano Mancini}
\author[iit]{Ferdinando Cannella}

\address[iit]{Istituto Italiano di Tecnologia, Via Morego 30, 16163 Genoa, Italy}
\address[dibris]{Department of Informatics, Bioengineering, Robotics and Systems
  Engineering (DIBRIS), University of Genoa, 16145 Genoa, Italy}
\address[ecampus]{Department of Theoretical and Applied Sciences (DISTA), eCampus
  University, Via Isimbardi 10, 22060 Novedrate (CO), Italy}
\address[karazin]{Department of Intelligent Software Systems and Technologies,
  School of Computer Science and Artificial Intelligence, V.\,N.\ Karazin Kharkiv
  National University, 4 Svobody Sq., 61022 Kharkiv, Ukraine}
\address[univpm]{Department of Information Engineering, Marche Polytechnic
  University, 60131 Ancona, Italy}

\begin{abstract}
Convolutional Kolmogorov--Arnold Networks (KANs) replace the fixed weights of a
convolutional kernel with learnable univariate functions. The dominant formulation
attaches one such function to every kernel entry and lets it act on pixel values,
expressive but parameter-heavy and prone to overfitting. We argue
that the learnable functions are better placed in the \emph{structure} of the
convolution than on each edge, and we organise the design space along a single axis:
whether the function acts on the pixel \emph{values} or on the filter \emph{shape}. We
study three realisations. SV-KAN applies one shared univariate function to the values
and leaves the spatial filter free and static, a classical convolution with a single learnable shared activation. AG-KAN keeps the shared
value function but supplies the spatial structure through a content-adaptive Gaussian
gate. RF-KAN instead moves the learnable functions onto the filter shape, building each
filter from oriented ridge profiles expanded in a localised oscillatory (Morlet)
wavelet basis with content-adaptive amplitudes. Under a matched four-layer protocol
with in-run references and three seeds, RF-KAN and SV-KAN reach $88.47\pm0.10\%$ and
$88.20\pm0.31\%$ on CIFAR-10 and $64.40\pm0.19\%$ and $64.57\pm0.30\%$ on CIFAR-100, at
about $0.4$M parameters. At this matched scale the shape model and the simplest value
model meet at the top, both above a plain convolution and every per-edge KAN we tested, including the official Gram variant, at roughly a fifth of the parameters. A controlled
study attributes the RF-KAN gain to an intrinsically localised oscillatory basis and to content adaptivity,
and an ablation that removes the learned shape entirely, leaving only the shared value
function, collapses accuracy by over forty points, identifying the learned shape as the
load-bearing ingredient at this scale.
\end{abstract}
 
 
\end{frontmatter}
 

\section{Introduction}
\label{sec:intro}
 
Kolmogorov--Arnold Networks~\cite{liu2024kan} place learnable univariate functions on
the edges of a network and sum them at the nodes, rather than learning linear weights
followed by a fixed activation. The construction is inspired by the Kolmogorov--Arnold
representation theorem, though no practical KAN realises it exactly. The variants that
have followed differ mainly in the basis for the univariate functions: B-splines in the
original work, radial basis functions in FastKAN~\cite{li2024fastkan}, wavelets in
Wav-KAN~\cite{bozorgasl2024wavkan}, and orthogonal polynomials such as Chebyshev,
Legendre and Gram.
 
A convolutional form is needed for vision. The established approach~\cite{bodner2024convkan},
extended with several bases by Drokin~\cite{drokin2024convkan}, replaces every kernel
weight with an independent univariate function of the corresponding input value, the
\emph{per-edge} formulation. It is expressive but costly: the number of functions scales
with input channels, output channels and kernel area, which inflates parameters and
inference time and tends to overfit. At a matched budget on natural images the evidence
is sober, with surveys and dedicated studies reporting that convolutional KANs match or
trail comparable networks while using more
resources~\cite{kansurvey,cankanwork2024,efficiency2025}; the Gram variant, reported as
the strongest~\cite{drokin2024convkan}, leads a plain convolution only at several times
its parameters. Strong results exist but are largely domain-specific, in medical
imaging~\cite{medkan,taylorkan2025}, remote
sensing~\cite{jamali2024hsi,hyperkan2024,cheon2024satellite} and continual
learning~\cite{cacciatore2024continual}. For ordinary image classification at a fixed
budget the picture is parity at best, which is the setting we address.
 
This paper starts from a simple observation: a patch carries the pixel \emph{values} and
their spatial \emph{positions}, and a learnable univariate function can act on either,
with very different inductive biases. Acting on the value is the per-edge primitive;
acting on the position defines the filter \emph{shape}, while the values enter linearly.
Rather than place one free function on every edge, we place a single shared function where
it buys the most, and organise three architectures along this value--shape axis. Two are
value KANs: SV-KAN (Shared-Value KAN) leaves the spatial filter free and static, a
classical convolution whose fixed activation is replaced by one shared learnable function,
while AG-KAN (Adaptive-Gate KAN) supplies the spatial structure through a content-adaptive
Gaussian gate. The third, RF-KAN (Ridge-Function KAN), places the learnable functions on
the filter shape, as oriented ridge profiles in a Morlet wavelet basis with
content-adaptive amplitudes. All three beat the per-edge KANs of the literature at a
matched budget, and RF-KAN and SV-KAN also beat a plain convolution of the same scale. We
characterise the source of the RF-KAN gain, separating the free basis from content
adaptivity, and map the surrounding design space with negative results, including an
ablation that removes the learned shape. All comparisons use a matched four-layer protocol
on CIFAR-10 and CIFAR-100 with identical training and in-run references.

\section{Related Work}
\label{sec:related}

\paragraph{Convolutional KANs (per-edge)}
ConvKAN~\cite{bodner2024convkan} introduced the convolutional KAN by replacing each
kernel weight with a spline-parametrised univariate function, and reported
competitive accuracy with up to half the parameters of a classical convolution on
Fashion-MNIST. Drokin~\cite{drokin2024convkan} extended the construction to a range
of bases (B-splines, radial basis functions, Chebyshev, Legendre and Gram
polynomials) and to ResNet- and DenseNet-like models, and proposed parameter-efficient
designs; among the bases the Gram variant performs best, although the models reported
to match or beat classical convolutions do so with several times more parameters and
more than twice the inference time. Memory and compute-efficiency of KAN training
has since become a research target in its own right~\cite{metakan2025}, and a natural
way to curb the per-edge cost is to \emph{share} the univariate function rather than
diversify it: Light-ResKAN~\cite{yi2026lightreskan}, for example, shares a Gram
activation across the positions of a channel and drops the learned spatial
aggregation, recovering spatial structure through the depth of a residual backbone for
SAR recognition. FastKAN~\cite{li2024fastkan} replaces splines with Gaussian radial
basis functions for efficiency, and Wav-KAN~\cite{bozorgasl2024wavkan} introduces
wavelet activations. These are all formulations in which the learnable function acts
on the pixel \emph{value}, differing in the basis used or in how the function is
shared, rather than in where the learnable structure is placed. Our RF-KAN instead
places a wavelet basis on the spatial \emph{coordinate} to define the filter shape:
Wav-KAN learns a nonlinear map of intensities, whereas RF-KAN learns the geometry of a
linear filter. Our ablations show that sharing a single value function is sufficient
(diversifying it per filter or channel does not help at a matched budget) but that
removing the learned shape collapses accuracy on a compact backbone, which is what
separates our shared-value member, SV-KAN, from a shape-free sharing scheme.
 
\paragraph{Content-adaptive and dynamic convolutions}
A separate line of work makes the convolution itself depend on the input. CondConv
and dynamic convolution mix a small bank of kernels per image with learned attention
weights, ODConv adds attention along several axes, and deformable convolutions adapt
the sampling positions. Most relevant to us, adaptive convolutions with per-pixel
dynamic atoms~\cite{wang2021acda} generate filters at each position by combining a
lightweight set of basis atoms whose coefficients are predicted from local features.
RF-KAN with content-adaptive amplitudes is related to this family. The difference is
that RF-KAN constrains the adaptive filter to a free univariate ridge expansion
rather than to generic atoms, which keeps it a KAN and makes its filters
interpretable.
 
\paragraph{Positional encoding in KAN convolutions}
The combination of positional information with a convolutional KAN has appeared
concurrently in PE-ConvKAN~\cite{peconvkan2026}, which embeds a fixed sinusoidal
absolute and relative positional encoding into the intermediate features of a
spline-based per-edge ConvKAN to preserve temporal order and phase in
one-dimensional vibration sequences for structural health monitoring. Our AG-KAN is
different in mechanism and purpose: the spatial component is not an additive
sinusoidal position vector but a learned, content-adaptive Gaussian gate that
multiplicatively reshapes the two-dimensional receptive field of each patch, and the
value path is a single shared function rather than per-edge splines. The two works
share only the high-level pairing of position with a convolutional KAN.
 
\paragraph{Ridge functions and the representation theorem}
A function of the form $\psi(\langle \vu, \vn\rangle)$ is a ridge function, constant
along the hyperplane orthogonal to $\vn$. Sums of ridge functions are the object of
projection pursuit and are universal approximators in the limit of many directions.
This is the structure that appears inside the Kolmogorov--Arnold representation, and
it is the one we use for the filter shape. Adaptive ridge directions have been used
inside KANs for function approximation, for example in active-subspace embedded
KANs~\cite{askan}, which identify dominant ridge directions for fitting scalar
functions; we instead use ridge profiles to parametrise convolutional filters in
vision.

\section{Method}
\label{sec:method}

\subsection{Spatial KAN convolution}
\label{sec:method:general}

Let a patch be $\mathcal{P}\in\mathbb{R}^{C\times P\times P}$, the $C$-channel
$P\times P$ window the convolution reads at one location, extracted with unit stride
and reflective padding. Spatial positions are written $\vu=(u,v)$ on a grid normalised
to $[-1,1]$, so that the centre of the patch is the origin and the corners are
$(\pm1,\pm1)$; this makes the coordinate the natural argument for the filter shape in
RF-KAN, independently of $P$. We cast the operators we study in a single form, which we call a \emph{spatial KAN convolution}: it applies 
$S$ filters and mixes their responses linearly into $C_{\mathrm{out}}$
 output channels. Writing $x_c(\vu)$
for the value at position $\vu$ in input channel $c$, the generic form is
\begin{equation}
  z_{s,c} \;=\; \sum_{\vu} w_s(\vu)\,\rho\big(x_c(\vu)\big),
  \qquad
  \mathrm{out} \;=\; \mathrm{BN}\big(W\,\mathbf{z}\big),
  \label{eq:generic}
\end{equation}
where $s\in\{1,\dots,S\}$ indexes the filter and $c$ the input channel; $w_s(\vu)$ is
the value of filter $s$ at position $\vu$; $\rho$ is a pointwise map applied to the
pixel values; and $z_{s,c}$ is the resulting scalar response of filter $s$ on channel
$c$ for the patch. Stacking all $S\!\cdot\!C$ responses into the vector $\mathbf{z}$,
a linear projection $W$ mixes them into the $C_{\mathrm{out}}$ output channels,
followed by batch normalisation, ReLU and average pooling.

\paragraph{Per-edge baseline}
The literature formulation removes $w_s$ and instead attaches an independent
univariate function $\phi_{s,c,\vu}$ to each kernel entry, acting on the value,
$z_{s} = \sum_{c,\vu}\phi_{s,c,\vu}\big(x_c(\vu)\big)$. With a basis of degree $D$
this introduces on the order of $C\,C_{\mathrm{out}}\,P^2\,(D{+}1)$ parameters per
layer. We implement Legendre and Chebyshev instances as matched baselines and use the
original implementation of the Gram variant.

\subsection{Learnable function placement}
\label{sec:method:placement}

Our models are separated by a single question: what is the argument of the learnable
univariate function. A patch carries the pixel \emph{values} and their spatial
\emph{positions}, and the function can act on either. This gives three families. The
\emph{per-edge} baseline attaches an independent function to every kernel entry, on
the value. The \emph{value} family keeps one shared function on the values and supplies
the spatial structure separately; its two members, AG-KAN and SV-KAN, differ only in
how that structure is produced. The \emph{shape} family, RF-KAN, places the learnable
functions on the filter shape itself, with the values entering linearly. The next two
subsections develop the value family and then the shape family; the formal consequence
of the distinction, that RF-KAN is linear in the patch values while the others are not,
is stated where it is used (Section~\ref{sec:method:rfkan}).

\subsection{Univariate functions}
\label{sec:method:psi}

All three families reuse a learnable univariate function written in residual form, so
that the layer starts from the identity and adds nonlinearity only where it helps:
\begin{equation}
  \psi(t) \;=\; t \;+\; \alpha \sum_{m=1}^{M} a_m\, \kappa\!\left(\frac{t-\mu_m}{\sigma_m}\right),
  \label{eq:psi}
\end{equation}
with fixed centres $\mu_m$, learnable amplitudes $a_m$, learnable widths $\sigma_m$,
and a scalar gate $\alpha$ initialised at zero. The kernel $\kappa$ is a Gaussian for
the radial-basis variant and a real Morlet wavelet,
$\kappa(z)=\cos(\omega_0 z)\,e^{-z^2/2}$, for the wavelet variant. The Morlet kernel
is localised and oscillatory in a single atom, a property that turns out to matter
(Section~\ref{sec:results}).

\subsection{KAN on the values: AG-KAN and SV-KAN}
\label{sec:method:value}

Both members apply the shared $\psi$ to the pixel values and map the result to the
output channels by the dense projection $W$ of Eq.~\eqref{eq:generic}. They differ only
in the form of $w_s$, the spatial structure applied before the projection: an adaptive
Gaussian gate in AG-KAN, a free static filter bank in SV-KAN. With the value
nonlinearity fixed and shared, the two ask how much of the spatial shape needs to be
learned, and how.

\paragraph{AG-KAN: adaptive Gaussian gate}
AG-KAN supplies the spatial structure through a content-adaptive Gaussian gate. A small
routing network reads the flattened patch and predicts the parameters
$(\sigma_u,\sigma_v,\theta)$ of a normalised anisotropic Gaussian on the patch grid,
\begin{equation}
  g(\vu)\;\propto\;\exp\!\Big(-\tfrac{u'^2}{2\sigma_u^2}-\tfrac{v'^2}{2\sigma_v^2}\Big),
  \qquad (u',v') = R_\theta\,\vu,
  \qquad \sum_{\vu} g(\vu) = 1,
  \label{eq:pe}
\end{equation}
and modulates the patch before the nonlinearity,
$\tilde{x}(\vu)=x(\vu)\,[1+\alpha_{\mathrm{pe}}(g(\vu)-1)]$. The gate lets the effective
receptive field change with local content; the Gaussian form is the minimal
interpretable parametrisation of such a field, a position with two scales and an
orientation, and it is stable in training where less constrained adaptive shapes
collapse (Section~\ref{sec:results}).

\paragraph{SV-KAN: free static shape}
SV-KAN replaces the gate with a free, static filter bank: the spatial weights
$w_s(\vu)$ are learnable parameters with no shape prior and no content adaptivity,
exactly as in a classical convolution, while the value path remains the shared $\psi$.
It is the minimal value KAN, a standard convolution whose fixed activation is replaced
by one learnable shared function on the values. It shows that the value paradigm
reaches the competitive regime in its simplest form, and it serves as the control that
isolates the learned shape: comparing it against variants that diversify $\psi$ per
filter or per channel, and against one that removes the learned shape entirely,
separates the effect of sharing from that of the shape (Section~\ref{sec:results}).

\subsection{KAN on the filter shape: RF-KAN}
\label{sec:method:rfkan}
 
\begin{figure*}[t]
\centering
\includegraphics[width=\textwidth]{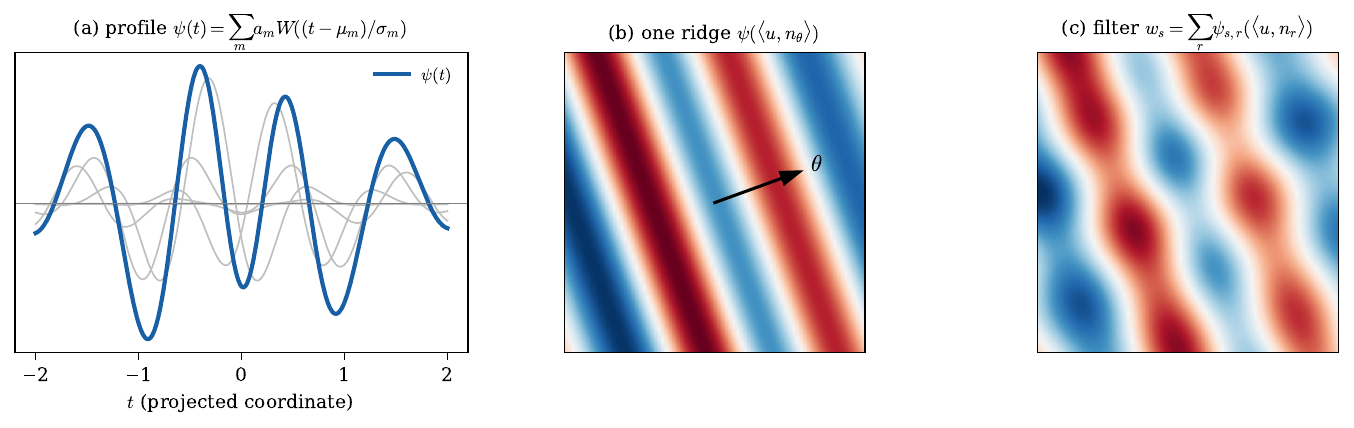}
\caption{\textbf{Construction of an RF-KAN filter.}
  (a) A free univariate profile $\psi(t)$ is a weighted sum of Morlet atoms (grey),
  each localised and oscillatory. (b) Every patch position $\vu$ is projected onto a
  learnable orientation $\theta$ to give a scalar $t=\langle\vu,\vn_\theta\rangle$,
  and the filter value at that position is $\psi(t)$. Positions with the same
  projection share a value, which produces an oriented ridge. (c) The sum of two
  ridges along different orientations yields a two-dimensional filter. The filter is
  a continuous field of signed values constrained to samples of a few smooth
  profiles, which is the source of both its regularity prior and its expressivity
  ceiling.}
\label{fig:construction}
\end{figure*}

\begin{figure}[t]
\centering
\includegraphics[width=\columnwidth]{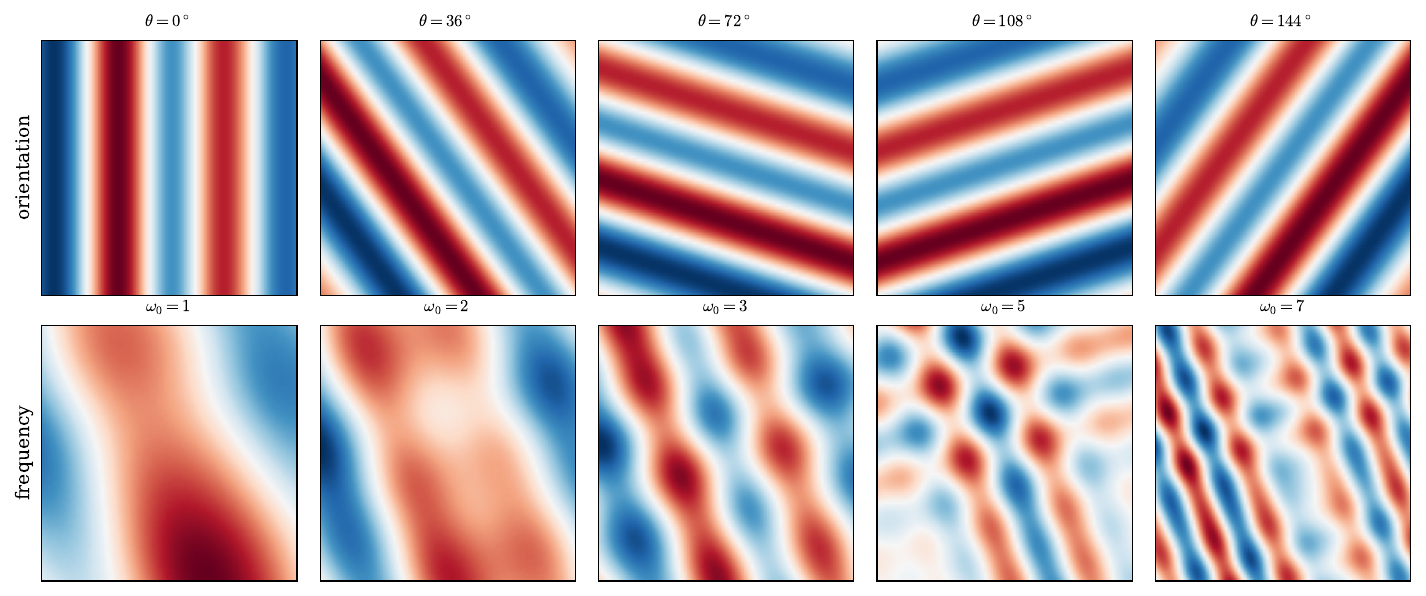}
\caption{\textbf{The filter family spanned by RF-KAN.}
  Top: a single ridge as the learnable orientation $\theta$ varies. Bottom: a
  two-ridge filter as the Morlet central frequency $\omega_0$ varies. Orientation and
  frequency are continuous and learned; the profiles themselves are free within the
  basis.}
\label{fig:gallery}
\end{figure}

In RF-KAN the learnable functions define the filter shape and the values enter
linearly. Each of the $S$ filters is a sum of $R$ ridge profiles taken along learnable
orientations,
\begin{equation}
  w_s(\vu)\;=\;\sum_{r=1}^{R}\psi_{s,r}\big(\langle \vu,\vn_{s,r}\rangle\big),
  \qquad \vn_{s,r}=(\cos\theta_{s,r},\sin\theta_{s,r}),
  \label{eq:ridge}
\end{equation}
where each profile $\psi_{s,r}$ is expanded in the Morlet basis of
Eq.~\eqref{eq:psi}. The argument $t=\langle\vu,\vn_{s,r}\rangle$ is a \emph{scalar}: the
two coordinates of $\vu$ and the orientation collapse, through the inner product, into a
single projected coordinate, so $\psi_{s,r}$ is univariate even though the filter it
produces is two-dimensional. All positions sharing the same projection receive the same
value, which makes $w_s$ constant along the direction orthogonal to $\vn_{s,r}$, a
ridge; the orientation $\theta_{s,r}$ is itself learnable, setting where the ridge
points while the profile sets what is detected along it. Because the values enter
Eq.~\eqref{eq:generic} linearly, RF-KAN is a KAN on the coordinates rather than the
values; doubling the patch values doubles its output, whereas in the value and per-edge
families it does not. We do not claim RF-KAN realises the representation theorem, whose
inner functions need not be smooth, whereas the Morlet expansion is smooth and finite;
what it provides is a controlled regularity bias, since the kernel entries are tied to
samples of a few smooth profiles, with the ridge count $R$ setting how much of the
$P\times P$ kernel space the filter can reach. Each profile is one continuous curve: the
$M$ Morlet atoms of Eq.~\eqref{eq:psi} overlap and combine additively into a single
smooth oscillatory shape set by the learnable amplitudes, sampled on a fine grid and
only then reduced to the small kernel, which is therefore a discretisation of a
continuous function rather than a set of free entries. Figure~\ref{fig:construction}
shows the construction and Figure~\ref{fig:gallery} the family of filters it spans as
orientation and central frequency vary.
 
\paragraph{Content-adaptive amplitudes}
The amplitudes are made content-adaptive: a routing network reads the channel-averaged
patch and predicts an additive correction to the base amplitudes,
$a_{s,r,m}=a^{\mathrm{base}}_{s,r,m}+\Delta a_{s,r,m}(\mathcal{P})$, so the filter shape
adapts to local content. The routing output is initialised near zero but not exactly
zero, which avoids a gradient dead-lock in which the branch never activates; we found
this necessary in practice.
 
\paragraph{Implementation}
The filter is continuous in the coordinates: we render it on a grid of size $Q$ and map
it back to the $P\times P$ kernel through a fixed bilinear matrix $U$,
$w^{\mathrm{eff}}_s = w_s U$. Because bilinear upsampling is linear, folding it into the
filter is exact and avoids interpolating the much larger field of stacked patches,
removing the dominant cost; the image stays discrete while the filter remains
continuous, and the aggregation in Eq.~\eqref{eq:generic} runs on the original $P^2$
positions.

\section{Experimental Setup}
\label{sec:exp}

All models share a four-layer convolutional backbone with channel widths
$3\!\to\!32\!\to\!64\!\to\!128\!\to\!256$, each layer followed by batch normalisation,
ReLU and average pooling, then global pooling and a linear classifier; we swap only the
spatial operator, so the comparison is about the operator rather than scale. Training
uses AdamW (learning rate $10^{-3}$, weight decay $10^{-4}$), batch size $128$, with
standard augmentation, for $50$ epochs on CIFAR-10 and $60$ on CIFAR-100. Every
experiment is repeated for three seeds. Because run-to-run variation from
non-deterministic GPU kernels is $0.1$ to $0.5$ points at this scale, every comparison
includes the same reference model \emph{within each run} and we do not compare across
runs. Baselines use the same hyperparameters as our models and are not retuned per
operator. We report the best test accuracy under a fixed, shared schedule, which leaves
the model-to-model comparison unaffected. All experiments ran on a single NVIDIA
GeForce RTX 3090 (24\,GB), PyTorch~2.10, CUDA~12.8.

For the per-edge baselines we implement Legendre (KALN) and Chebyshev (KACN)
convolutions in the same backbone, with a base path on $\mathrm{SiLU}(x)$ added to a
polynomial path on the basis expansion of the tanh-normalised input. For the Gram
variant (KAGN), the strongest per-edge KAN, we use the original
implementation~\cite{drokin2024convkan} unchanged, forcing its internal normalisation
to batch normalisation so that the comparison isolates the basis rather than the
normalisation; with its native instance normalisation the same layer scores about four
points lower, and its internal activation remains SiLU. As a non-KAN reference we train
a canonical Gabor filter (an anisotropic Gaussian envelope times an oriented carrier,
in the sense of Daugman) in both a static and a content-adaptive variant whose routing
matches our models.

\section{Results}
\label{sec:results}

\subsection{Main comparison}

Table~\ref{tab:main} reports CIFAR-10 accuracy, parameter counts and the difference
against the plain convolution. RF-KAN reaches $88.47\pm0.10\%$ at $0.40$M parameters,
$1.87$ points above a plain convolution of the same scale and above every per-edge
convolutional KAN. SV-KAN, the shared-value member with a free static shape, follows
closely at $88.20\pm0.31\%$ and the same $0.39$M budget, so the value paradigm reaches
the competitive regime in its simplest form; AG-KAN, the adaptive-gate value model,
reaches $86.87\%$. The strongest per-edge baseline, the Gram variant from its original
implementation, reaches $85.95\%$, marginally above Legendre ($85.39\%$) and Chebyshev
($83.70\%$) and consistent with the literature ranking, yet it trails RF-KAN by about
$2.5$ points while using roughly five times the parameters. A canonical Gabor filter
with the same content-adaptive routing as RF-KAN reaches only $85.77\%$, and $84.46\%$
without adaptivity, so RF-KAN also clears a standard parametric filter of the same
adaptive kind. Every structural KAN we propose exceeds all per-edge baselines; RF-KAN
and SV-KAN additionally exceed the plain convolution.

\begin{table}[H]
\centering\small
\setlength{\tabcolsep}{5pt}
\caption{CIFAR-10, four-layer backbone, three seeds. Parameter counts in millions.
  Our structural KAN convolutions in \textbf{bold}. The last column is the accuracy
  difference against the plain convolution. Per-edge KANs and the Gabor baselines are
  trained in the same backbone; the Gram variant uses its original implementation.}
\label{tab:main}
\begin{tabular}{llrr}
\toprule
Model & Accuracy (\%) & Params (M) & $\Delta$ CNN \\
\midrule
\textbf{RF-KAN (wavelet, adaptive)}   & $\mathbf{88.47 \pm 0.10}$ & \textbf{0.40} & $\mathbf{+1.87}$ \\
\textbf{SV-KAN (shared value, free shape)} & $\mathbf{88.20 \pm 0.31}$ & \textbf{0.39} & $\mathbf{+1.60}$ \\
\textbf{RF-KAN (wavelet, static)}     & $\mathbf{87.97 \pm 0.10}$ & \textbf{0.39} & $\mathbf{+1.37}$ \\
\textbf{Dual KAN (value + ridge)}     & $\mathbf{87.17 \pm 0.27}$ & \textbf{0.39} & $\mathbf{+0.57}$ \\
\textbf{AG-KAN (shared value + gate)} & $\mathbf{86.87 \pm 0.17}$ & \textbf{0.43} & $\mathbf{+0.27}$ \\
Plain convolution                     & $86.60 \pm 0.17$ & 0.39 & $0.00$ \\
\textbf{RF-KAN (RBF ridge, static)}   & $\mathbf{86.17 \pm 0.57}$ & \textbf{0.39} & $\mathbf{-0.43}$ \\
Per-edge KAN, Gram (official)         & $85.95 \pm 0.05$ & 1.94 & $-0.65$ \\
Adaptive Gabor (aniso., non-KAN)      & $85.77 \pm 0.39$ & 0.40 & $-0.83$ \\
Per-edge KAN, Legendre                & $85.39 \pm 0.41$ & 1.94 & $-1.21$ \\
Gabor (aniso., static, non-KAN)       & $84.46 \pm 0.15$ & 0.39 & $-2.14$ \\
Per-edge KAN, Chebyshev               & $83.70 \pm 0.34$ & 1.94 & $-2.90$ \\
\midrule
\multicolumn{4}{l}{\emph{Diagnostic ablation: shared value, learned shape removed}} \\
SV-KAN, no shape (uniform sum) & $46.14 \pm 0.12$ & 0.39 & $-40.46$ \\
\bottomrule
\end{tabular}
\end{table}

\begin{table}[H]
\centering\small
\setlength{\tabcolsep}{5pt}
\caption{CIFAR-100, four-layer backbone, three seeds. Parameter counts in millions.
  Our structural KAN convolutions in \textbf{bold}. The last column is the accuracy
  difference against the plain convolution. For compactness we report only the most
  significant models per category; per-edge KANs and the Gabor baseline are trained in
  the same backbone.}
\label{tab:cifar100}
\begin{tabular}{llrr}
\toprule
Model & Accuracy (\%) & Params (M) & $\Delta$ CNN \\
\midrule
\textbf{SV-KAN (shared value, free shape)} & $\mathbf{64.57 \pm 0.30}$ & \textbf{0.39} & $\mathbf{+2.90}$ \\
\textbf{RF-KAN (wavelet, adaptive)}   & $\mathbf{64.40 \pm 0.19}$ & \textbf{0.42} & $\mathbf{+2.73}$ \\
\textbf{AG-KAN (shared value + gate)} & $\mathbf{62.59 \pm 0.21}$ & \textbf{0.45} & $\mathbf{+0.92}$ \\
Plain convolution                     & $61.67 \pm 0.48$ & 0.41 & $0.00$ \\
Adaptive Gabor (aniso., non-KAN)      & $61.09 \pm 0.41$ & 0.42 & $-0.58$ \\
Per-edge KAN, Gram (official)         & $60.43 \pm 0.13$ & 1.97 & $-1.24$ \\
Per-edge KAN, Legendre                & $58.20 \pm 0.63$ & 1.97 & $-3.47$ \\
Per-edge KAN, Chebyshev               & $55.92 \pm 0.26$ & 1.97 & $-5.75$ \\
\bottomrule
\end{tabular}
\end{table}

On CIFAR-100 the ordering holds and the margins widen (Table~\ref{tab:cifar100}).
SV-KAN and RF-KAN lead at $64.57\pm0.30\%$ and $64.40\pm0.19\%$, within noise of each
other and about $2.8$ points above the plain convolution; the strongest per-edge KAN
(Gram, $60.43\%$) trails by roughly four points and the adaptive Gabor ($61.09\%$) by
more than three, the per-edge baselines again at five times the parameters. That the
shape paradigm (RF-KAN) and the value paradigm (SV-KAN) meet at the top, from opposite
ends of the value--shape axis, is the main empirical message: both placements of the
learnable function reach the same competitive regime at a matched budget, and neither
needs the per-edge multiplication of functions.
Read against the per-edge baselines, the result is a statement about accuracy per
parameter: our structural designs reach this regime at a matched budget, whereas the
per-edge formulation attains comparable accuracy only at several times the parameter
count, and it does so in precisely the compact, natural-image setting where
convolutional KANs have been least convincing.

\subsection{Ablation and design boundaries}
\label{sec:results:ablation}

\paragraph{Sources of the RF-KAN gain}
The advantage of RF-KAN separates into two largely independent contributions, each
visible as a drop when removed. Replacing the Morlet basis with a radial-basis ridge
lowers accuracy from $87.97\%$ to $86.17\%$, about $1.3$ points at the tuned central
frequency; removing content adaptivity from the full model lowers it from $88.47\%$ to
$87.97\%$, a further $0.8$ points. A routing gain of comparable size appears
independently for the radial-basis ridge and for a static Gabor filter (about $1.3$
points), so adaptivity is a general lever rather than an artefact of one basis. The
central frequency is the most influential hyperparameter: the static wavelet ridge
scores $87.97\%$, $87.45\%$ and $86.87\%$ at $\omega_0=3,5,7$, while the atom count
saturates early ($M=6,8,12,16$ all within noise), so we fix $M=6$.

\paragraph{Free basis versus parametric template}
To separate the free basis from adaptive geometry, we compared against a canonical
Gabor filter under identical routing. With the same adaptivity the Gabor reaches
$85.8\%$ on CIFAR-10 and $61.1\%$ on CIFAR-100, trailing RF-KAN by about $2.7$ and
$3.5$ points, and its anisotropic envelope is within noise of the isotropic one. The fixed template we tested thus falls short of the free Morlet ridge even when granted the same
adaptivity.

\paragraph{Sharing and geometry}
Within SV-KAN, sharing a single value function, assigning one per filter, and assigning
one per channel are equivalent in accuracy ($88.20$, $87.89$, $87.92\%$ on CIFAR-10, all
within noise). Since diversifying the function brings no gain while adding parameters,
we share it; this parallels the classical convolution, which shares an activation across
filters and diversifies through the linear weights. The learned shape, by contrast, is
not optional: replacing it with a fixed uniform sum over positions, the same operator
otherwise, collapses CIFAR-10 accuracy to about $46\%$ (Table~\ref{tab:main}), a drop of
more than forty points.

\section{Conclusions}
\label{sec:conclusion}

We presented structural Kolmogorov--Arnold convolutions that place the learnable
univariate function in the structure of the convolution rather than on every edge, along
a value--shape axis. RF-KAN places it on the filter shape, as ridge profiles in a Morlet
wavelet basis with content-adaptive amplitudes; SV-KAN and AG-KAN keep a single shared
function on the values, with a free static shape and an adaptive gate respectively. At a
matched four-layer scale all three beat the per-edge convolutional KANs of the
literature, including the Gram variant in its original implementation, at about a fifth
of the parameters, and RF-KAN and SV-KAN also beat a plain convolution of equal size,
reaching $88.5$ and $88.2\%$ on CIFAR-10 and $64.4$ and $64.6\%$ on CIFAR-100. An important finding is that the
most expressive shape model and the simplest value model coincide in accuracy: at this scale the placement of the learnable function matters more than
its multiplicity.

Beyond accuracy, the structural placement may aid interpretability, which we offer as a
hypothesis: where a per-edge KAN exposes one plottable function per kernel entry, on the
order of $C\,C_{\mathrm{out}}\,P^2$ per layer, AG-KAN reduces this to a single shared
function plus a gate whose orientation and two scales read off directly. Evaluation
beyond the present compact backbone is underway: extensions to Fashion-MNIST and Tiny
ImageNet, to deeper residual backbones and to ImageNet scale, together with the effect
of the ridge count $R$, characterisation of the additional latent capacity of RF-KAN
that we expect to separate it from SV-KAN at larger scale, a rank-controlled measurement
of the expressivity ceiling, and a full accounting of FLOPs and latency, are the subject
of ongoing work.

\section*{Data and Code Availability}
Implementation and training scripts will be released on acceptance.
\bibliographystyle{elsarticle-num}

\bibliography{references}

\end{document}